\documentclass{svproc}
\usepackage{url}
\usepackage{booktabs}

\usepackage{multirow}
\usepackage{float}
\usepackage{graphicx} %package to manage images
\graphicspath{ {./images/} }
\usepackage[symbol]{footmisc}

\usepackage{lipsum}

\begin{document}

\mainmatter              % start of a contribution
\title{On Sensitivity of Deep Learning Based Text Classification Algorithms to Practical Input Perturbations}
\titlerunning{Sensitivity of Deep Learning Models to Input Perturbations} 
\author{Aamir Miyajiwala  \inst{1,3,}\thanks{All authors contributed equally} \and Arnav Ladkat \inst{ 1,3,*} \and
Samiksha Jagadale  \inst{1,3,*} \and Raviraj Joshi\inst{2,3}}

\authorrunning{Aamir Miyajiwala et al.} % abbreviated author list (for running head)

%%% list of authors for the TOC (use if author list has to be modified)
% \tocauthor{Ivar Ekeland, Roger Temam, Jeffrey Dean, David Grove,
% Craig Chambers, Kim B. Bruce, and Elisa Bertino}

\institute{Pune Institute of Computer Technology, Pune, Maharashtra, India \and
Indian Institute of Technology Madras, Chennai,  Tamilnadu, India
\and
L3Cube, Pune, Maharashtra, India \\
\email{\{aamir.miyajiwala, arnav.ladkat, samiksha0321\}@gmail.com}
\email{ravirajoshi@gmail.com}
}

\maketitle              % typeset the title of the contribution

\begin{abstract}
% The abstract should summarize the contents of the paper
% using at least 70 and at most 150 words. It will be set in 9-point
% font size and be inset 1.0 cm from the right and left margins.
% There will be two blank lines before and after the Abstract. \dots

Text classification is a fundamental Natural Language Processing task that has a wide variety of applications, where deep learning approaches have produced state-of-the-art results. While these models have been heavily criticized for their black-box nature, their robustness to slight perturbations in input text has been a matter of concern. In this work, we carry out a data-focused study evaluating the impact of systematic practical perturbations on the performance of the deep learning based text classification models like CNN, LSTM, and BERT-based algorithms. The perturbations are induced by the addition and removal of unwanted tokens like punctuation and stop-words that are minimally associated with the final performance of the model. We show that these deep learning approaches including BERT are sensitive to such legitimate input perturbations on four standard benchmark datasets SST2, TREC-6, BBC News, and tweet\_eval. We observe that BERT is more susceptible to the removal of tokens as compared to the addition of tokens. Moreover, LSTM is slightly more sensitive to input perturbations as compared to CNN based model. The work also serves as a practical guide to assessing the impact of discrepancies in train-test conditions on the final performance of models.

% We would like to encourage you to list your keywords within
% the abstract section using the \keywords{...} command.
\end{abstract}

\section{Introduction}

Text classification is a fundamental task of NLP and is used in simplifying tasks like document
classification, sentiment analysis of consumer feedback, and fake news detection \cite{kowsari2019text}\cite{kulkarni2021l3cubemahasent}\cite{wani2021evaluating}. Deep learning models have been used widely to tackle text classification as they tend to perform best compared to
other machine learning techniques \cite{wagh2021comparative}\cite{kulkarni2021experimental}. Some of the most popular methods used for this task are
based on convolutional neural networks (CNNs) and recurrent neural networks (RNNs) \cite{zhang2015sensitivity}\cite{zhou2015c}\cite{joshi2019deep}.
Pre-trained language models such as Bidirectional Encoder Representations from Transformers (BERT)
based on transformer architecture, perform better than neural networks trained from scratch \cite{devlin2018bert}.
Although the accuracy of the model has always been the primary focus of evaluation, it might
overestimate the performance of NLP models and also, fails to account for the robustness of the model to
small changes in the input data \cite{ribeiro2020beyond}\cite{liu2021robustness}\cite{goel2021robustness}\cite{kitada2021attention}. The sensitivity of deep learning models to input data leads to practical challenges while deploying the stochastic models in a consumer environment. 
It is therefore important to study the sensitivity of the models to small input perturbations.  In this work, we aim to highlight the limitations of deep learning based text classification models in different practical scenarios \cite{paleyes2020challenges}\cite{jacovi2020towards}. The analysis also highlights the need to keep deterministic checks in place while building real-time machine learning systems.  

Previous research concerning the robustness of models, studied the
vulnerabilities of DNNs by devising adversarial attacks \cite{singh2020model}\cite{la2020assessing}\cite{jin2020bert}\cite{gao2018black}\cite{nguyen2015deep}. These attacks were
carried out by making small perturbations in the input sample, primarily in the form of a word
substitutions \cite{moradi2021evaluating}. Our work, however, is not concerned with attacks on DNN or random input perturbations. In this work, we take a black-box approach to determine the impact on the performance of DNNs by inducing input perturbations in the form of systematic addition and removal of tokens. A similar black-box study evaluating the impact of word order on model performance was carried out in \cite{taware2021shuftext}. We use standard pre-processing methods as a proxy to introduce simple valid perturbations. Our experimental setup can be described as the application of standard pre-processing on training data followed by evaluation of the performance on raw un-processed text data and vice-versa. We create discrepancies in the train and test data by introducing practical changes like the presence or absence of punctuation, stop-words, and out-of-vocabulary words. We show that deep learning based models are sensitive to these simple addition or removal of unwanted tokens. We alternate between trainable and static word embeddings during the training to evaluate their impact. The three broad classes of models based on CNN, LSTM, and Transformer-based BERT are evaluated on standard SST2, TREC, BBC News, and tweet\_eval datasets. The motivation behind this approach is that these conditions represent a real-world environment where one could unknowingly have such discrepancies. Also, this study is a step forward in determining if any particular type of input configuration gives the best performance across any model. 

The paper is organized as follows. We present an overview of the effects of various pre-processing techniques and noisy data on text classification in Section \ref{sec:relatedWork}. Our experimental setup where we discuss the datasets, models, methodology, and proposed approach, is mentioned in Section \ref{sec:experimentalSetup}. Finally, the results and interpretations are summarized in Section \ref{sec:results} with the conclusion in Section \ref{sec:conclusion}.

\section{Related Work}
 \label{sec:relatedWork}

Our work is at the intersection of pre-processing methods and evaluating model robustness. This section briefly describes the previous works related to pre-processing techniques, noise present in the data, and their individual effects on the text classification models. Various studies on the pre-processing techniques can be seen in the literature, with the aim to find the most optimal combination of cleaning methods. More than 26 pre-processing methods were identified in \cite{albalawi2021investigating} and were applied to the data from Arabic-language social media. The results showed that a specific combination of four out of 26 techniques, the removal of duplicate words, and three other variants of normalizing Arabic letters, improved the F1 score of the MNB classifier by two percent.

A similar experiment was done in \cite{hacohen2020influence}, where they explored the impact of six basic pre-processing methods i.e. spelling correction, upper to lower case letters, reduction of repeated characters, HTML tag removal, stopwords removal, and punctuation removal, on four benchmark text corpora using three machine learning methods. It was seen that at least one combination of basic pre-processing methods could be recommended to improve the accuracy results.

An extensive comparison of fifteen pre-processing methods on CNN, LSTM, and BERT in \cite{effrosynidis2017comparison} across two Twitter datasets namely SS-Twitter and the SemEval, concludes that stemming, replacement of repetitions of punctuation, and removing numbers are the recommended techniques to be applied. The non-recommended techniques comprised of spelling correction, replacing negations with antonyms, handling capitalized words, replacing slang, and removing punctuation.

A study of stopwords removal on the sentiment classification models done in \cite{ghag2015comparative}. They found that based on term weighing techniques, the traditional sentiment classifier displays a 9 percent rise inaccuracy when stopwords are removed. Other approaches like ARTFSC, SentiTFIDF, and RTFSC, did not vary much on the removal of stopwords.

\cite{kreek2018training} compares the accuracy of the clean dataset with cross-validation performance measured on the dirty training dataset by listing the performance of CNN, fastText, and bag-of-words SVM performance on 20-Newsgroups, 2016 Yelp Reviews, and a synthetically created dataset from five different document collections. It was observed that the clean dataset continuously outperformed the cross-validation results on the dirty training dataset. However, they do not report results on experiments in which there is a combination of clean and unclean datasets for the training and testing data.

Sensitivity analysis of one-layer CNN's was performed in \cite{zhang2015sensitivity}. They report the results of experiments exploring different configurations of CNN's run over nine sentence classification datasets. It displays the effects of input word vectors, filter region size, the number of feature maps for each filter region size, activation function, pooling strategy, and regularisation on the one-layer CNN model. This study focuses on the model hyper-parameters whereas we look at the sensitivity from the input data perspective.

Our work is on the lines of \cite{moradi2021evaluating}. However, instead of using random character and world-level perturbations, we use input perturbations that are induced by the addition and removal of tokens in the form of stopwords and punctuations.

\section{Experimental Setup}
 \label{sec:experimentalSetup}

\subsection{Dataset description}
\label{sect:pdf}

We use Stanford Sentiment Treebank (SST-2)\cite{socher2013recursive}, Text Retrieval Conference (TREC-6)\cite{voorhees2000overview}, TweetEval\cite{barbieri2020tweeteval} and BBC News\footnote[1]{http://mlg.ucd.ie/datasets/bbc.html} datasets for our study. These datasets cover both binary and multi-class classification. The Stanford Sentiment Treebank consists of sentences from movie reviews and human annotations of their sentiment. It uses the two-way (positive/negative) class split, with only sentence-level labels, and consists of a total of 8741 data samples, out of which 6920 are training set samples and 1821 are test set samples with an average of 19 tokens in a sentence. The TREC-6 dataset consists of open-domain, fact-based questions categorized into 6 classes: Abbreviation, Description and abstract concepts, Entities, Human beings, Locations, and Numeric values. This dataset has 5452 training samples and 500 test samples with an average of 10 tokens in a sentence. The TweetEval dataset consists of seven heterogeneous tasks on Twitter Data, all framed as multi-class tweet classification. The tasks include seven tasks namely - irony, hate, offensive, stance, emoji, emotion, and sentiment. We perform Text Classification on the sentiment analysis task. It consists of 45615 sentences in the training data, 12284 in the test data, and 200 in the validation set. The data is classified into three labels. The BBC News dataset consists of 2225 documents from the BBC news website, having stories that belong to five topical areas of business, entertainment, politics, sport, and tech.

\subsection{Models}
\label{ssec:layout}

We use the most basic architectures from CNN and LSTM families in order to avoid any bias towards architecture-specific modifications.
The CNN model uses two convolutional layers with 512 and 256 filters in the first and second layer, each having a filter size of 5. This is followed by a GlobalMaxPool Layer. Two linear layers are stacked on top of the GlobalMaxPool Layer, one having 128 nodes and the other being the output layer. A Dropout of 20\% is also included before the output layer. The second classifier is a uni-directional LSTM model having two LSTM layers with 512 nodes each, followed by a GlobalMaxPool Layer. Two linear layers are further stacked on the GlobalMaxPool layer, one containing 256 nodes and the other being the output layer.

The third classifier we have used is the BERT base cased\footnote[2]{https://huggingface.co/docs/transformers/model\_doc/bert} model, pre-trained on the BookCorpus (800M words) and English Wikipedia  (2,500M words) dataset \cite{wolf2020transformers}. It consists of 12 layers of bidirectional Transformer based encoder blocks, wherein each layer has a total of 12 self-attention heads \cite{devlin2018bert}.
We observe that the BERT base uncased model provides similar results so we only focus on the cased model to avoid repetition. 
%We also used the BERT base uncased model for our work but since it showed similar behavior as the cased model we have not added its details.

\subsection{Methodology}

The methodology includes pre-processing methods, training of the word embeddings, and the use of OOV tokens. Each of the following has been described in detail below.

\subsubsection{Pre-processing techniques}

For cleaning of the dataset, we have used the subsequent pre-processing techniques:
\begin{enumerate}
\item \textbf{Expanding contractions.} Contractions are words or combinations of words that are shortened by dropping letters and replacing them with an apostrophe. Here, we are removing such contractions and replacing them with expanded words. For example, contractions like `can't' and `mightn't' are expanded to `can not' and `might not', respectively.
\item \textbf{Removal of special characters}. Non-word characters i.e. all the characters which are not from [a to z], [A to Z], and [0 to 9] are removed. Thus, we are getting rid of characters like `\verb|\|' and `!' from the data. 
\item \textbf{Removal of single characters.} After the removal of special characters from `\verb|\t|' and `\verb|\n|', single characters like `t' and `n' are left behind. For the removal of such single characters, this technique of pre-processing is necessary.
\item \textbf{Substituting multiple spaces with single space.} Due to the application of all the above techniques, a lot of multiple spaces tend to be generated. For the removal of these unnecessary spaces, we substitute those multiple spaces with a single space.
\item \textbf{Lowercasing.} The most common pre-processing technique is to convert all the words to lowercase. Suppose the word `Bold' occurs twice, once as `BOLD' and the second time as `Bold'. Both of them are represented as two different words in the vector space model, even though they mean the same. To avoid such unnecessary vectors and reduce the dimensions, lowercasing is used.
\item \textbf{Removal of Stopwords.} Stopwords are a collection of most frequently occurring words like `an', `of', `the'. Such words do not reflect a huge impact on the model since they have the least information required for the training of the model. By removing these stopwords, we tend to shift the focus from less important words to more important words, thus improving the accuracy of the classification models.
\end{enumerate}

The first five steps are grouped together and termed basic pre-processing steps. Removal of stop-words is separately considered during the analysis since it leads to large-scale token removal or addition.
We have used the data in different configurations as described below:
\begin{itemize}
    \item \textbf{Clean Data:} Apply the first five basic pre-processing steps to the data.
    \item \textbf{Clean Data - stop words:} Further remove stop words as well after applying basic pre-processing.
    \item \textbf{Unclean Data:} No pre-processing is applied on this data and non-alphanumeric tokens are separated from main tokens with space.
    % \item \textbf{Unclean Data - stop words:} The stopwords are removed from the unprocessed data.
\end{itemize}
These pre-processing steps are used to simulate the addition and removal of tokens during model evaluation. The addition of tokens can be simulated by training the model is on pre-processed (clean) data and evaluating it on the un-processed (unclean) data. Similarly, the removal of tokens can be simulated by training the model on unclean data and evaluating it on clean data. These simple pre-processing steps that affect only the unwanted tokens can be used to induce perturbations for evaluating the sensitivity of models to input tokens. Since we are not affecting the important tokens the drop inaccuracy of the model will signify the sensitiveness of the model to input tokens.   

\subsubsection{Word Embeddings}
In this work, we are using pre-trained FastText word embeddings that are an extension of the popular Word2Vec embeddings. The FastText word embedding does not consider the vectors for words directly, each word is represented as n-gram of characters. The benefit of using FasText over Glove or traditional Word2Vec is that it will be able to generate an embedding even for an out of vocab word as it splits the word up into n-gram of characters, which can be found in the training data.

The embedding layer of CNN and LSTM models is initialized using pre-trained fast text word vectors. The embeddings layer can be kept `trainable' to finetune its weights to work with the downstream task. In our analysis, we have experimented with both trainable and static word embeddings.

\subsubsection{OOV Tokens}
Word-based text processing models suffer from the problem of the Out Of Vocabulary(OOV) token due to the limited size of the vocabulary. A special OOV token is reserved for the out of vocab words encountered during testing. We drop 10\% of the least frequent words in the training data from the vocabulary so that model sees the OOV token during training as well. During testing or evaluation, a model may encounter OOV tokens in practice. For our study, we evaluate the effect of OOV tokens on the model performance. The input perturbations are simulated in such a way that it results in the addition of OOV tokens. To simulate this the model is trained on pre-processed clean data and the vocabulary is created using tokens remaining post-cleanup. The unwanted or filtered tokens are not made part of the vocabulary. So now when the model is tested on unclean data with all the tokens the unwanted tokens are mapped to OOV token thus simulating the addition of OOV tokens.

\begin{figure*}[ht]
    \centering
    \includegraphics[scale=0.43]{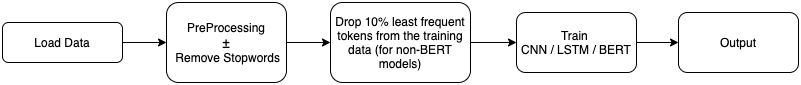}
    \caption{Proposed Approach}
    \label{fig:approach}
\end{figure*}

\begin{figure*}[ht]
    \centering
    \includegraphics[scale=0.58]{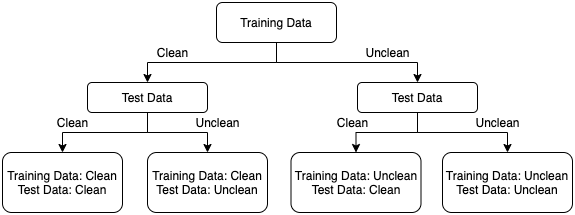}
    \caption{Prepared Data Sets}
    \label{fig:datasets_1}
\end{figure*}

\subsection{Proposed Approach}
We assess the robustness of the model by simulating perturbations in the form of discrepancies between train and test data. The configurations under consideration are shown in Figure \ref{fig:datasets_1}. Only one set is pre-processed at a time thus leading to training test mismatch. These discrepancies can be segregated into 3 ways:
\begin{itemize}
\item \textbf{Addition of tokens}: Here we train on clean data and test on unclean data. The vocabulary is created before any pre-processing of the data to retain all the tokens. The extra tokens in unclean data will not be mapped to OOV in this case. Note that the vocabulary and OOV token specifications are not relevant to the BERT model.
\item \textbf{Addition of OOV tokens}: Here we train on clean data and test on unclean data. However, the vocabulary is created using the pre-processed data to leave out unwanted tokens from the vocabulary. Thus the extra tokens in unclean data will now map to the OOV token as they are not part of the vocabulary.
\item \textbf{Removal of tokens}: This corresponds to training on unclean data and testing on clean data. The vocabulary is created using unclean data.
\end{itemize}
The clean data mentioned in the above three configurations is pre-processed in two ways:
\begin{itemize}
\item Basic pre-processing which involves the first 5 steps of pre-processing defined previously.
\item Basic pre-processing with the removal of stopwords. Removal of stop words helps in large-scale addition and removal of tokens as compared to only employing basic pre-processing.
\end{itemize}
A comparative analysis is performed with respect to the above configurations and their impact on the performance of models has been discussed in the next section.

\begin{table}[hbt!]
\centering
\begin{tabular}{@{}llllll@{}}
\toprule
\textbf{Dataset} & \textbf{Config} & \textbf{OT} & \textbf{TT} & \textbf{OT{[}RS{]}} & \textbf{TT{[}RS{]}} \\ \toprule
SST -2   & C     & 1,998  & 30,277   & 1,995  & 17,629   \\
         & C-U   & 3,038  & 32,891   & 17,132 & 32,891   \\
         & C-2   & 2,018  & 30,277   & 2,005  & 17,629   \\
         & C-U 2 & 2,071  & 32,891   & 2,071  & 32,891   \\
         & U     & 2,031  & 32,891   & 2,031  & 32,891   \\
         & U-C   & 2,006  & 30,277   & 2,005  & 17,629   \\ \toprule
TREC-6   & C     & 349    & 3,123    & 349    & 1,493    \\ 
         & C-U   & 443    & 3,309    & 443    & 3,309    \\
         & C-2   & 350    & 3,123    & 350    & 1,493    \\
         & C-U 2 & 357    & 3,309    & 357    & 3,309    \\
         & U     & 357    & 3,309    & 357    & 3,309    \\
         & U-C   & 350    & 3,123    & 350    & 1,493    \\ \toprule
Tweet Eval       & C               & 16,053      & 179,281    & 16,051              & 118,452            \\ 
         & C-U   & 26,988 & 193,589 & 92,961 & 193,589 \\
         & C-2   & 16,139 & 179,385 & 16,139 & 118,452 \\
         & C-U 2 & 19,255 & 193,489 & 19,255 & 193,589 \\
         & U     & 16,091 & 179,381 & 16,091 & 118,452 \\
         & U-C   & 19,205 & 193,589 & 19,205 & 193,589 \\ \toprule
BBC News & C     & 4,273  & 170,845 & 4,269  & 100,423 \\ 
         & C-U   & 11,316 & 183,377 & 87,433 & 183,377 \\
         & C-2   & 4,574  & 170,854 & 4,574  & 100,423 \\
         & C-U 2 & 4,738  & 183,377 & 4738   & 183,377 \\
         & U     & 4,531  & 179,854 & 4531   & 1,00,423 \\
         & U-C   & 4,493  & 183,377 & 4493   & 183,377 \\ \toprule
\end{tabular}
\caption{Number of Unknown and Total Token in all four Datasets; Total Tokens  (TT); OOV Token - Unknown Token (OT) in \%; RS (Removed Stopwords); C (Clean); C-U (CleanUnclean); U (Unclean); U-C (UncleanClean); C 2 (Clean, Generation of Vocabulary before pre-processing); C-U 2 (CleanUnclean, Generation of Vocabulary before pre-processing)}
\label{tab:token-countTable}
\end{table}

\begin{table*}[hbt!]
\centering
\begin{tabular}{lllllllllll}
\hline
\textbf{Model}           & \textbf{EmTr} & \textbf{Dataset} & \multicolumn{2}{l}{\textbf{SST-2}} & \multicolumn{2}{l}{\textbf{Trec-6}} & \multicolumn{2}{l}{\textbf{tweet-eval}} & \multicolumn{2}{l}{\textbf{BBC News}} \\ \cline{4-11} 
     &       &       & \textbf{Acc}   & \textbf{Acc[RS]} & \textbf{Acc}   & \textbf{Acc{[}RS{]}} & \textbf{Acc}   & \textbf{Acc{[}RS{]}} & \textbf{Acc}     & \textbf{Acc{[}RS{]}} \\ \hline
CNN  & False & C     & 71.49 & 69.30       & 83.16 & 62.96       & 61.09 & 59.68       & 95.37 & 97.35     \\
     &       & C-U   & 69.20 & 56.42       & 81.20 & 53.08       & 59.73 & 57.83       & 93.53 & 86.83     \\
     &       & C 2   & 74.67 & 69.18       & 83.60 & 60.88       & 61.30 & 60.45       & 95.51 & 96.22     \\
     &       & C-U 2 & 73.47 & 69.52       & 83.56 & 57.00       & 59.43 & 58.63       & 94.83 & 95.01     \\
     &       & U     & 73.26 & 71.95       & 83.76 & 84.28       & 60.65 & 61.04       & 94.83 & 94.97     \\
     &       & U-C   & 72.27 & 70.33       & 84.76 & 41.56       & 60.84 & 58.74       & 93.30 & 93.66     \\ \cline{2-11} 
     & True  & C     & 77.19 & 76.17       & 88.24 & 71.88       & 59.09 & 59.11       & 95.28 & 97.35     \\
     &       & C-U   & 76.27 & 59.78       & 84.00 & 61.04       & 58.62 & 59.33       & 94.61 & 91.69     \\
     &       & C 2   & 77.90 & 75.89       & 86.96 & 74.00       & 59.57 & 58.93       & 95.51 & 96.58     \\
     &       & C-U 2 & 77.17 & 75.58       & 87.28 & 55.88       & 59.67 & 59.13       & 95.28 & 95.15     \\
     &       & U     & 77.05 & 77.40       & 86.48 & 86.44       & 60.49 & 59.72       & 95.19 & 95.51     \\
     &       & U-C   & 76.80 & 74.66       & 88.60 & 46.40       & 60.27 & 58.78       & 93.93 & 93.57     \\ \hline
LSTM & False & C     & 76.34 & 72.62       & 82.84 & 72.52       & 64.19 & 63.62       & 95.60&	95.73     \\
     &       & C-U   & 71.30 & 51.73       & 82.44 & 45.20       & 63.76 & 62.56       & 91.51&	65.98     \\
     &       & C 2   & 77.02 & 74.49       & 83.12 & 69.84       & 64.01 & 62.48       &94.74&	94.79     \\
     &       & C-U 2 & 75.29 & 74.06       & 81.72 & 60.48       & 64.78 & 63.59       &93.89&	88.54     \\
     &       & U     & 76.83 & 74.50       & 83.48 & 85.08       & 64.58 & 64.75       & 93.21&	94.47     \\
     &       & U-C   & 74.94 & 72.50       & 83.60 & 63.64       & 64.40 & 62.67       & 95.46&	94.88     \\ \cline{2-11} 
     & True  & C     & 77.47 & 75.57       & 84.92 & 73.00       & 61.25 & 59.29       & 96.81&	96.81     \\
     &       & C-U   & 77.94 & 63.21       & 85.16 & 56.92       & 60.59 & 58.37       & 94.29&	72.90     \\
     &       & C 2   & 78.90 & 77.40       & 85.40 & 73.40       & 61.65 & 59.74       & 94.25&	95.33     \\
     &       & C-U 2 & 78.77 & 75.98       & 85.60 & 57.52       & 60.54 & 58.94       & 94.16&	92.72     \\
     &       & U     & 78.30 & 77.91       & 86.36 & 85.96       & 61.22 & 60.75       & 95.42&	95.91     \\
     &       & U-C   & 76.86 & 74.71       & 85.56 & 69.16       & 61.03 & 59.84       & 93.98&	95.42     \\ \hline
BERT-base & True & C       & 90.62       & 84.11       & 96.16        & 85.64       & 67.19          & 64.97         & 97.66       & 97.66      \\
     &       & C-U   & 90.73 & 86.82       & 86.44 & 55.20       & 67.44 & 65.42       & 98.43 & 97.98     \\
     &       & U     & 91.11 & 90.94       & 97.24 & 96.88       & 68.26 & 68.13       & 98.02 & 98.16     \\
     &       & U-C   & 90.01 & 81.79       & 93.00 & 48.80       & 67.14 & 64.29       & 94.97 & 92.94    \\ \hline
\end{tabular}
\caption{Results for all four Datasets; EmTr (Embeddings Trainable); Acc (Accuracy) in \%; RS (Removed Stopwords); C (Clean); C-U (CleanUnclean); U (Unclean); U-C (UncleanClean); C 2 (Clean, Generation of Vocabulary before pre-processing); C-U 2 (CleanUnclean, Generation of Vocabulary before pre-processing)}
\label{tab:combined-table}
\end{table*}

\section{Results}
\label{sec:results}
In this section, we discuss the sensitivity of our models with respect to the defined perturbations. The accuracy of the model are compared with clean(train) - clean(test) and unclean(train) - unclean(test) baselines. In the baseline models, both train and test data follow the same pre-processing steps. The results for the four datasets are described in Table \ref{tab:combined-table}. The statistics of out of vocabulary tokens for each configuration are given in Table \ref{tab:token-countTable}. For the TREC dataset, all the models are very sensitive to the input perturbations with a significant drop in performance when we remove stopwords during preprocessing. We performed an analysis on the number of tokens after pre-processing with the removal of stopwords, against the raw text. In the case of raw text, we have an average of 10 tokens in the data, but when we remove stopwords during preprocessing, only two tokens remain in the TREC-6 dataset. This attributes to the extreme sensitivity in the performance of all the models on TREC-6 data.

\begin{figure*}[hbt!]
    \centering
    \includegraphics[scale=0.5]{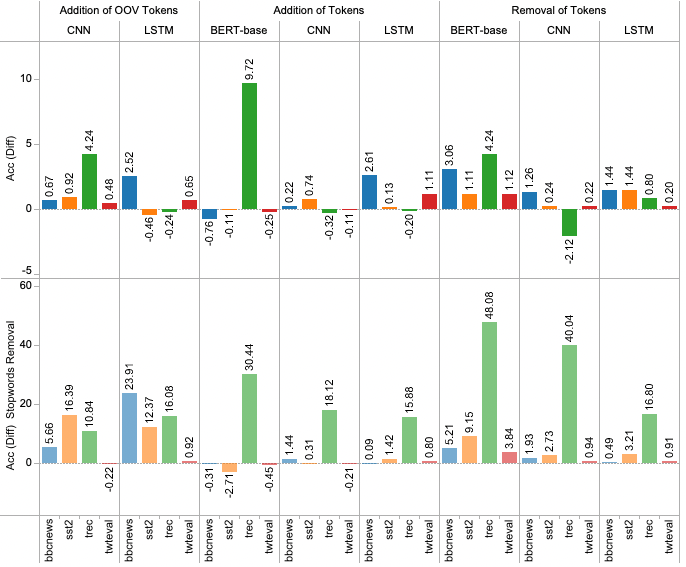}
    \caption{The accuracy difference for trainable token embeddings configuration, without and with Stopwords Removal; Addition of OOV Tokens consists of the difference in the accuracy of Clean (C) and accuracy of Clean-Unclean (C-U); Addition of Tokens consists of the difference in the accuracy of Clean 2 (C 2) and accuracy of CleanUnclean 2 (C-U 2); Removal of Tokens consists of the difference in the accuracy of Unclean (U) and accuracy of UncleanClean (U-C)}
    \label{fig:TRUE}
\end{figure*}

\begin{figure*}[hbt!]
    \centering
    \includegraphics[scale=0.5]{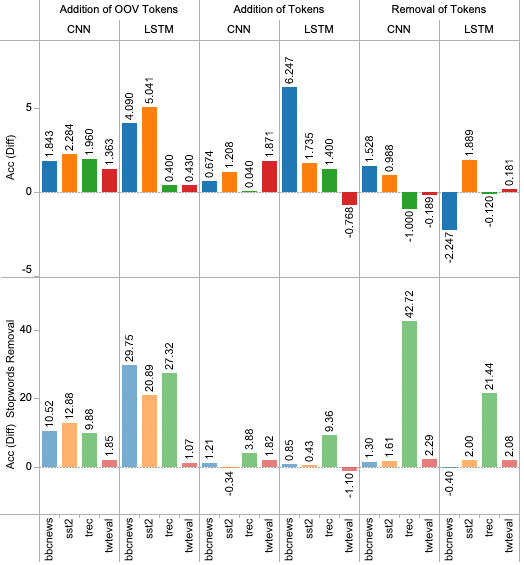}
    \caption{The accuracy difference for static token embeddings (trainable=False) configuration, without and with Stopwords Removal; Addition of OOV Tokens consists of the difference in the accuracy of Clean (C) and accuracy of Clean-Unclean (C-U); Addition of Tokens consists of the difference in the accuracy of Clean 2 (C 2) and accuracy of CleanUnclean 2 (C-U 2); Removal of Tokens consists of the difference in the accuracy of Unclean (U) and accuracy of UncleanClean (U-C)}
    \label{fig:FALSE}
\end{figure*}

\subsection{Addition of Tokens}
The addition of tokens corresponds to the clean-unclean (C-U 2) case and is compared against the clean-clean (C 2) model as visualized in Figure \ref{fig:TRUE} and Figure \ref{fig:FALSE}. We notice that the addition of tokens doesn't have much effect on the CNN model and BERT model. There is a slight decrease in accuracy for CNN, however, the decrease is more pronounced for the LSTM model. Also, the impact of addition is more when the embeddings are static. Similarly, the addition of a large number of tokens in the form of stop words has more impact on the LSTM model.

\subsection{Removal of Tokens}
The removal of tokens corresponds to the unclean-clean (U-C) case and is compared against the unclean-unclean (U) model as shown in Figure \ref{fig:TRUE} and Figure \ref{fig:FALSE}.
The CNN model is found to be quite robust to the removal of small and large number tokens. Whereas the LSTM model shows a drop in performance even when a small number of tokens are removed for a couple of datasets. Finally in the case of the BERT model removing a small number of tokens doesn't have much impact but when a large number of tokens are removed even BERT shows a significant drop in performance. Since BERT is originally a language model such a drop is justified.

\subsection{Addition of OOV Tokens}
The addition of OOV tokens corresponds to the clean-unclean (C-U) case and is compared against the clean-clean (C) model in Figure \ref{fig:TRUE} and Figure \ref{fig:FALSE}. Also, this case is specific to CNN and LSTM models as there is no OOV token in the BERT model.
Both the models are robust to small additions in OOV tokens. However as we increase the number of OOV tokens, in terms of stop words, the performance drop for both CNN and LSTM models is significant. Again the drop in performance is more pronounced when the embeddings are static.

\section{Conclusion}
\label{sec:conclusion}

In this work, we analyze the impact of practical input perturbations on deep learning based text classification algorithms. The perturbations studied are in the form of the addition and removal of unwanted tokens. The unwanted tokens are characterized as punctuation and stop words which are minimally important for the classification process. The models under consideration are CNN, LSTM, and BERT. We show that these models are robust to a small number of additions or removals. However, as we increase the number of perturbations the performance of models drops significantly even for the BERT model. The models are more sensitive to removals as compared to additions. Also with trainable word embeddings, the models are more robust. We also find that CNN is more robust to such changes as compared to the LSTM model. Overall we show that these deep learning models are sensitive to such simple perturbations in input data.

\section*{Acknowledgements}
This work was done under the L3Cube Pune mentorship program. We would like to express our gratitude towards our mentors at L3Cube for their continuous support and encouragement.

\bibliographystyle{splncs03_unsrt}

\bibliography{main}

\end{document}